*Article*

# Spatiotemporal modeling of grip forces captures proficiency in manual robot control

**Rongrong Liu[1], John Wandeto[2], Florent Nageotte[3], Philippe Zanne[3], Michel de Mathelin[3] and Birgitta Dresp-Langley[3]***

[1] University of Strasbourg IDEX Postdoctoral Program, France; rongrong.liu.pro@gmail.com;
[2] Department of Information Technology, Dedan Kimathi University of Technology, Nyeri, Kenya; ndetos@gmail.com;
[3] ICube UMR 7357 Robotics Department University of Strasbourg, France; nageotte@unistra.fr; philippe.zanne@unistra.fr; demathelin@unistra.fr;
[4] ICube UMR 7357 University of Strasbourg, Centre National de la Recherche Scientifique (CNRS), France; birgitta.dresp@cnrs.fr;

*Correspondence: birgitta.dresp@cnrs.fr;

**Abstract:** New technologies for monitoring grip forces during hand and finger movements in non-standard task contexts have provided unprecedented functional insights into somatosensory cognition. Somatosensory cognition is the basis of our ability to manipulate and transform objects of the physical world and to grasp them with the right amount of force. In previous work, the wireless tracking of grip force signals recorded from biosensors in the palm of the human hand has permitted us to unravel some of the functional synergies that underlies perceptual and motor learning under conditions of non-standard and essentially unreliable sensory input. This paper builds on this previous work and discusses further, functionally motivated, analyses of individual grip force data in manual robot control. Grip forces were recorded from various *loci* in the dominant and non-dominant hands of individuals by means of wearable wireless sensor technology. Statistical analyses bring to the fore skill-specific temporal variations in thousands of grip forces of a complete novice and a highly proficient expert in manual robot control. A brain inspired neural network model that uses the output metric of a Self-organizing Map with unsupervised winner-take-all learning was run on the sensor output from both hands of each user. The neural network metric expresses the difference between an input representation and its model representation at any given moment in time *t* and reliably captures the differences between novice and expert performance in terms of grip force variability.Functionally motivated spatiotemporal analysis of individual average grip forces, computed for time windows of constant size in the output of a restricted amount of task-relevant sensors in the dominant (preferred) hand, reveal finger-specific synergies reflecting robotic task skill. The analyses lead the way towards grip force monitoring in real time. This will permit tracking task skill evolution in trainees, or identify individual proficiency levels in human robot-interaction, which represents unprecedented challenges for perceptual and motor adaptation in environmental contexts of high sensory uncertainty. Cross-disciplinary insights from systems neuroscience and cognitive behavioral science, and the predictive modeling of operator skills by parsimonious Artificial Intelligence (AI) will contribute towards improving the outcome of new types of surgery, in particular the single-port approaches such as NOTES (Natural Orifice Transluminal Endoscopic Surgery) and SILS (Single Incision Laparoscopic Surgery).

## 1. Introduction

Analysis of grip force signals tailored to hand and finger movement evolution for grip force control during task execution provides insight into the fundamental mechanisms of somatosensory cognition [1]. Recent technology has permitted the wireless monitoring of grip force signals recorded from biosensors in the palm of the human hand to track and trace human grip forces deployed in image-guided precision tasks under conditions of restricted sensory input [2,3]. Such grip force sensing permits profiling operator strategies and, at the same time, exploring functional interactions between somatosensory and motor control during the strategic planning and execution of hand movements, with a potential for generating benchmarks for human-robot interaction [4-6]. Somatosensory cognition is the basis of our ability to manipulate and transform physical objects [1,7], to recognize them on the basis of touch alone [8,9], and to grasp them with the right amount of force for lifting, manipulation, or transformation [10-14]. Sensorial and cognitive processes underlying hand-specific grip force variations (dominant *versus* non-dominant hand) in manual tasks have been studied in a variety of contexts by selectively probing multiple measurement locations in the fingers and palm of the human hand [12-17]. Grip force modulation is governed by neuronal connections that are potentiated on the basis of self-organized learning [18,19,20], which drives the development of functionally specific neural networks in the continuously learning brain. The neural activities in these networks are modulated by sensory signals processed in the somatosensory cortex, the so-called S1 map [21]. S1 corresponds to a neocortical area that responds primarily to tactile stimulation on the skin or hair and plays a critical role in grip force control in interaction with multiple sensory areas. Somatosensory neurons have the smallest receptive fields, receiving the shortest-latency input from their peripheral receptors. Their cortical functional organization is conceptualized in current state of the art [22-24] in terms of a single neural network map of the receptor periphery, with a modular functional architecture and highly specific connectivity patterns, coordinating functionally distinct neuronal subpopulations from other cortical areas involved in sensory processing into motor circuit modules at several hierarchical levels [23-27]. These functionally specific modules display a hierarchy of interleaved circuitry connecting via inter-neurons in the spinal cord, in the visual, auditory, and olfactory sensory areas and in the motor cortex, with feed-back loops and bilateral communication with the supraspinal centers [21,22]. Anatomically adjacent to motor cortex, S1 is thus functionally connected to all sensory areas [28,29] responding to stimuli from the environment (Fig. 1).

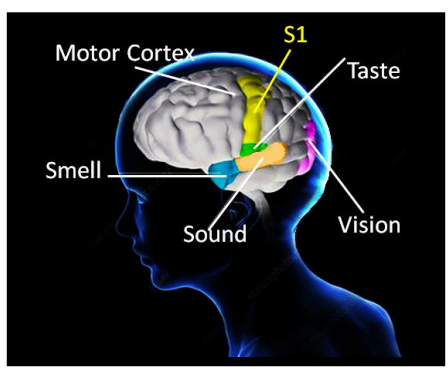

**Figure 1.** Anatomically adjacent to the motor cortex, the somatosensory brain (S1) controls mechanoreceptors and plays a critical role in grip force production through functional interactions with multiple sensory areas. The prefrontal lobe controls the conscious modulation of grip forces and motor behavior (not illustrated here).

Somatosensory afferents reach the frontal lobe and feed into circuitry for prefrontal responses to somatosensation (touch) and the conscious control of motor and grip force behaviors [30,31]. There is a strong functional link between visual and somatosensory cognition in sighted individuals, who have learnt to rely on visual input for motor planning and control, as visual input permits anticipate grip forces required for manipulating objects [32]. When visual cues are suddenly no longer available and the subject is confronted with high uncertainty about digit-force related object properties, sensorimotor memories take over [9,33]. Such memories form during life-long brain learning on the basis of previous grasp experience to permit cognitively controlled grip force adaptation when physical object properties and behavioral consequences cannot be reliably sensed. The contribution of each finger to overall grip strength, coarse adjustments, and finer grip force control vary across cognitive tasks and their diverse requirements for motor planning and execution. While the middle finger is critical for lifting and manipulating heavy objects in three dimensions [34], the ring finger and the small finger mostly control fine grip force modulation [35,36,37,38,39,40,41]. Subtle grip force deployment with minimal variation is necessary for precision tasks such as surgery. In precision tasks, the contribution of the index finger to total grip force is often the smallest, and there seem to be no significant differences between men and women, yet, the amount of force applied by each digit depends on several factors including where the digits are placed when grasping [34,42]. In the absence of external constraints, the complex anatomy of the human hand allows for a large variety of postures and force combinations to attain stable grips, generating functional synergies that permit solving the problem of motor redundancy [40,41,42,43,44]. Multi-finger grip force control relies on the self-organizing principles of from-local-to-global functional interaction and multiple feed-back loops at several hierarchical stages, from hand to brain and back [42,45,46]. The adaptive scaling of both magnitude and rate of hand or finger force is skill-specific. In the true expert, such scaling relies on high-level cognitive control mechanisms, finely tailored to the skill in question [47,48]. Expert surgeons, for example, not only deploy grip forces more parsimoniously than novices [49], and their spatiotemporal grip force profiles reveal patterns characteristic of expertise by comparison with novices or trainee surgeons [40,41]. Robot-assisted surgical training illustrates novel perspectives offered by modern grip force sensor technology for the study of functionally significant changes during task skill acquisition. Hand and finger grip forces directly impact on the trajectory and velocity of surgical tool displacements. Optimal hand grip force produces optimal object displacement trajectories and movement [10, 11]. Minimally invasive robotic surgery is an image-guided high-precision task where absence of haptic force feedback spontaneously yields stronger, often excessive (i.e. non-optimal) hand grip forces, especially in novices [50]. This can result in unnecessary or excessive tissue damage in a patient, and novices therefore have to learn to overcome this problem by scaling their finger forces accordingly. Such learning requires adaptation to unusual constraints, because robot-assisted surgical systems impose conditions of limited Degrees of Freedom (DoF) for hand and finger movements during the manipulation of the surgical tools attached to the system. The tools cannot straightforwardly be moved in any direction as in traditional surgery. This represents a considerable constraint for motor planning and control to which novices need to adapt. Also, veridical information about real-world depth is missing from the image representations on the screen of such surgical systems, and instead of looking down on his/her hands, the surgeon only sees the tool-ends controlled by the system. Camera and image calibration problems added, the tool movements displayed on a screen may not match the real-world movements in time and space, and the combined loss of real-world depth input and veridical space scale information significantly affects the performance of novices and experts who are not familiar with the system [37,38,51,52,53]. Robot assisted surgery thus profoundly challenges solidly formed perceptual, somatosensory, and cognitive representations of space, scale, and relative distance for eye-hand coordination [54,55]. Here in this work, we exploit thousands of individual grip force data recorded from wireless sensors in functionally relevant locations of the dominant hands of an expert and a novice to account for the evolution of each individual's motor behaviour during training in a four step pick-and-drop simulation task on a robotic platform designed for single-access transluminal endoscopic surgery.

## 2. Materials and Methods

A wireless wearable (glove) sensor system was used for collecting thousands of grip force data per sensor location and individual user in real time. The glove system does not provide haptic user feedback, and the sensors therein were positioned to fit the cylindrical shape of the control handles of a robotic system designed for bi-manual intervention in transluminal endoscopic surgery. Task simulations may solicit either the dominant, the non-dominant, or both hands at the same time depending on the complexity of the task. Here, grip force data recorded from the dominant and non-dominant hands of an expert and a novice tested in ten successive sessions, performed without breaks between sessions, of a four-step *pick-and-drop* task were generated. All sensors of the glove system were carefully calibrated, as explained in further detail here below in section *2.2*. The calibration results are displayed in the Results sections.

*2.1 Robotic System*

The robotic system is built on the Anubis® platform of Karl Storz. The slave system consists of three flexible, cable-driven sub-units for robot-assisted endoscopic surgery with ten motorized DoF, which are described in further detail in previous work [38]. The main endoscope carries a fisheye camera at its end, providing visual feedback. The lateral and longitudinal control of the endoscope is user-dependent, and the endoscope can be bent in two orthogonal directions, moving the endoscopic view from left to right, and up and down at any given position. The distal instruments at the tool-end are inserted in the channels of the endoscope, and they have bending extremities. This system has a tree-like architecture, where movements of the endoscope, controlled by the user via two cylindrically shaped handles, impact also upon the position and orientation of the distal extremities (tool-tips) (Fig. 2). Each of these has three DoF, two for translation and rotation in the endoscope channel, and one for the deflection of the active distal extremity/tool-tip. Translation, rotation, and deflection are electrically actuated (motorized) via cables running through the endoscope, from its proximal part to the distal end. The distal extremities/tool-tips open and close mechanically when a trigger is pulled by the user. The slave robot is controlled by a position loop running at 1000 Hz on the CPU of the master system, which consists of two specially designed interfaces, which are passive mobile mechanical systems. The user is to put his hands tightly around each of the two handles, each of which has three DoF for controlling the tool tips' translational and rotational movements, and those around the axis of movement of the initial DoF for the lateral or longitudinal control of the endoscope via a given handle, which the user manipulates with the left or the right hand depending on the task. The master interfaces are statically balanced and all joints exhibit low friction, which entails that only minimal forces are required to produce movement in any direction.

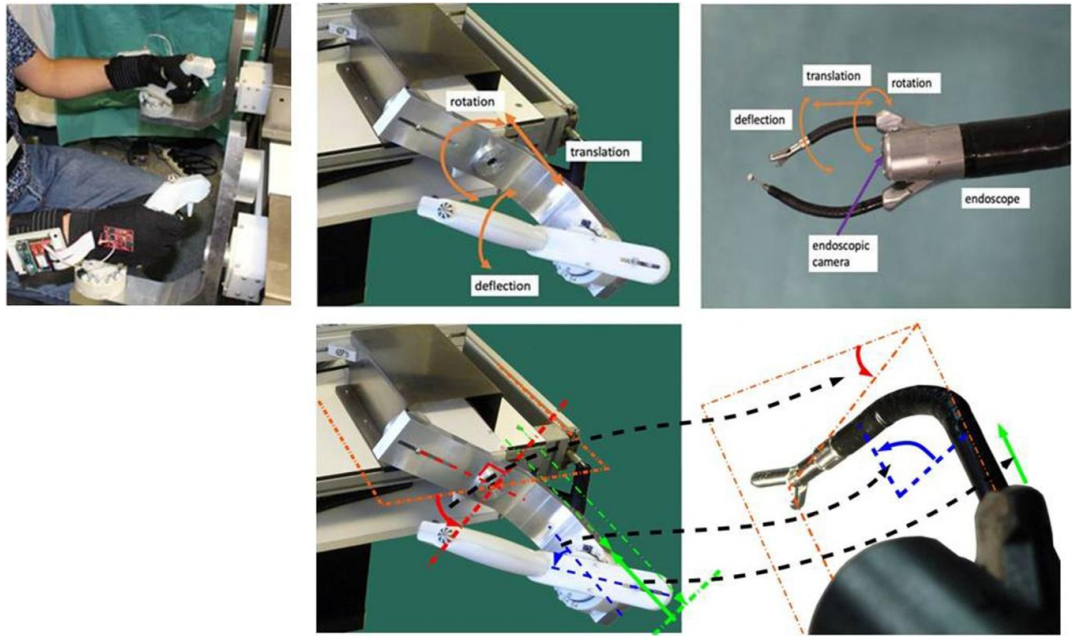

**Figure 2.** The sensor gloves in action (top left) on the handles of the robot master system. Joints of the master control system handles and directions of movement (top middle). Slave instruments at the distal extremities of the endoscope and directions of movement (top right). Master-slave movement mapping (bottom right).

The controller of the robot master system runs on a DELL Precision T5810 model computer equipped with an Intel Xeon CPU E5-1620 with 16 Giga bytes memory (RAM) under real-time Linux. The real-time Linux mapping control software communicates with the master interface and the slave system.

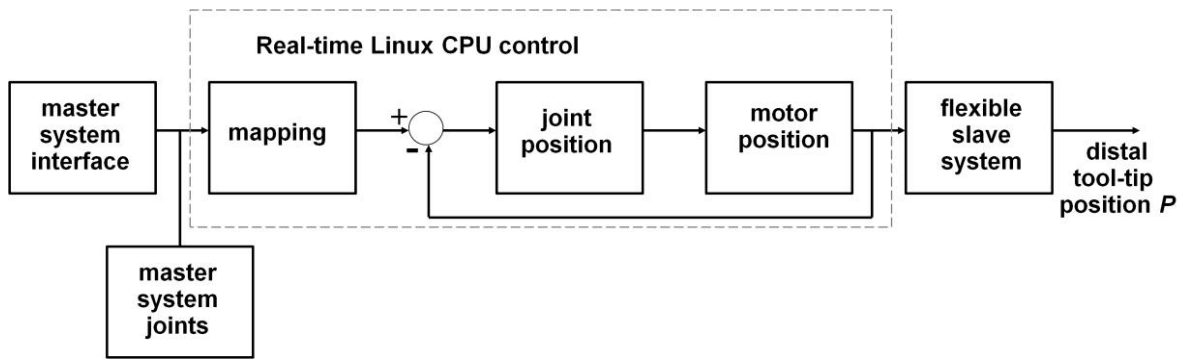

**Figure 3.** Master-slave control flow chart of the robotic system. In the 'mapping' block, the horizontal rotation of the master system handle is mapped to the rotation of the follower, the translation of the master is mapped to that of the follower, and the vertical rotation of the master is mapped to the bending angle of the follower as shown at the bottom of Figure 2. The mapping scales are indicated in the text here below.

Joint positions of the master interfaces are obtained from encoders read at 1kHz by the central controller of the master CPU, which maps the positions individually to the corresponding joints of the slave systems to compute reference positions. The master CPU sends these reference positions to the drivers controlling the slave motors at 1kHz. The slave joints are individually servoed to their reference positions by their drivers. The master-slave control flow chart of the robotic system is illustrated schematically here above (Fig. 3). The mapping scales from master to slave are 1:1 for rotations, 1:2 for bending, and 1:2:1 for translations. Each of the two handles has a trigger for controlling the mechanical opening and closing of the grippers at the tool-tips, and a small four-way joystick for controlling specific camera movements not required for the study (**NB** this joystick was not used in any of the experimental sessions here). In the experimental task, the trigger controlling the tool-tip graspers was operated with the index finger of the hand in action. The user sits in front of the master console and looks at the endoscopic camera view displayed on the screen in front of him/her at a distance of about 80 cm while holding the two master handles, which are about 50 cm away from each other. Seat and screen heights are adjustable to optimal individual comfort. Left and right master interfaces are identical and the two slave instruments they control are also identical. Therefore, for a given task the same movements need to be produced by the user whatever the hand he/she uses (left or right). A snapshot view of a user wearing the sensor gloves while manipulating the handles of the system is shown above (Fig. 2 top left).

*2.2 Sensor gloves*

A wearable sensor system in terms of two gloves, one for each hand, with inbuilt Force Sensitive Resistors (FSR) was developed. Each of the two gloves has 12 FSR that are in contact with specific locations on the inner surface of the hand, as illustrated here below (Fig. 4). The sensor locations were chosen for optimal contact with the surface of the cylindrically shaped robotic handles when holding them with each hand. The FSR were inserted between two layers of cloth and did not interact directly with ether the skin of the subject, or the surface of the master handles, which provided a comfortable feel during manipulations. FSR were sewn with needle and thread into the cloth layer around their conducting surfaces (active areas). The electrical connections of the sensors were individually routed to the dorsal side of the hand and brought to a soft ribbon cable, connected to a small and very light electrical casing, strapped onto the upper part of the forearm and equipped with an Arduino microcontroller. Eight of the FSR, positioned in the palm of the hand and on the finger tips, had a 10 mm diameter, while the remaining four located on middle phalanxes had a 5mm diameter. Each FSR was soldered to 10KΩ pull-down resistors to create a voltage divider, and the voltage read by the analog input of the Arduino is given by

$$V_{out} = R_{PD}V_{3.3}/(R_{PD}+R_{FSR}) \qquad (1)$$

where $R_{PD}$ is the resistance of the pull down resistor, $R_{FSR}$ is the FSR resistance, and $V_{3.3}$ is the 3.3 V supply voltage. FSR resistances can vary from 250Ω when subject to 20 Newton (N) to more than 10MΩ when no force is applied at all. The generated voltage varies monotonically between 0 and 3.22 Volt, as a function of the force applied, which is assumed uniform on the sensor surface. In the experimental task here, forces applied did not exceed 1100 gram (g), which corresponds to ~10 Newton. The relation between force and voltage is linear within the range of output voltages measured in the experiments here for variations within the range of [0; 1500] mV. Careful calibration of all sensors was performed prior to the experiments to ensure that all sensors provided similar calibration curves. The relationship between force (g) and tension (mV) here is shown graphically in the Results section. Regulated 3.3V was provided to the sensors from the Arduino. Power was provided by a 4.2V Li-Po battery enabling use of the glove system without any cable connections. The battery voltage level was controlled during the whole duration of the experiments by the Arduino electronic prototyping platform and displayed continuously via the user interface. The glove system was connected to a computer for data storage via Bluetooth enabled wireless communication at a rate of 115200 bits-per-second (bps).

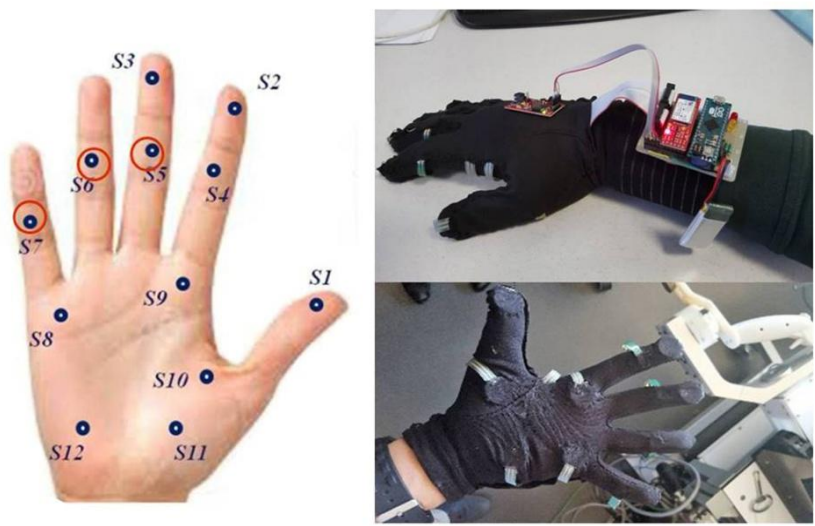

**Figure 4.** The sensor glove system has twelve sensors positioned mirror-symmetrically across the left and right hands. The spatiotemporal grip force profiles for expert and novice described under 3.3 in the Results section here were drawn from a functionally relevant subset of three sensors (red circles) in the dominant hand, translating optimal grip force deployment by the pinky (S7), and minimized gross grip force deployment by the ring and middle fingers (S5, S6) characteristic of expertise in controlling the master-slave system of the surgical robot. The data from ten of the twelve sensors (S2, S3, S5, S6, S7, S8, S9, S10, S11, S12) in the dominant and non-dominant hands of the two users were fed into the variability (standard deviations) and neural network analyses (SOM). Sensors S4 and S1 produced too little significant output and were not taken into account.

*2.3. Software*

The software of the glove system has two parts: one running on the gloves, and one for data collection. Each of the two gloves is sending data to the computer separately, and the software reads the input values, and stores them on the computer according to their header values indicating their origin. The software is running on Arduino and designed to acquire analog voltages provided by each FSR every 20 milliseconds (50Hz). In every loop, input voltages were merged with their time stamps and sensor identification. This data package was sent to the computer via Bluetooth, which was decoded by the computer software. The voltage data were saved in a text file for each sensor, with their time stamps and identifications. Furthermore, the computer software monitored the voltage values received from the gloves via a user interface showing the battery level. In case the battery level drops below 3.7 V, the system warns the user to change or charge the battery. However, this never occurred during the experiments reported here.

*2.4 Experimental precision grip task*

A *4-step pick-and-drop* precision grip task requiring specific device movements in all directions (left, right, forwards and backwards) had to be performed as swiftly and accurately as possible. A verbal description for each task step, is provided here below (Table 1). Visual illustrations are shown further below (Fig. 5).

**Table 1.** Verbal description for each of the four task steps. The colored boxes correspond to task steps as visualized in the spatiotemporal profiles in Figure 10 in the Results section.

| Task Step | Hand-Tool Interaction Required |
|---|---|
| 1 | Activate and move tool forwards towards the pick-up target box |
| 2 | Move tool downwards towards object, open grippers, close grippers on object, lift |
| 3 | Move tool in lateral direction towards the destination box for dropping object |
| 4 | Open grippers to drop object in the destination box |

During the experiments, only one of the two instruments controlling the tool-tips (left or right, depending on the task session) was moved, while the main endoscope and the camera image remained still. The experimental task starts with the right or left hand gripper being pulled back. Then the user has to approach the object by a forward movement in depth (*step 1*) of the distal tool extremity by manipulating the handles of the master system effectively. This forward movement in depth is the most difficult to perform under the conditions given (2D image guidance) because veridical depth information is unavailable [51,55].

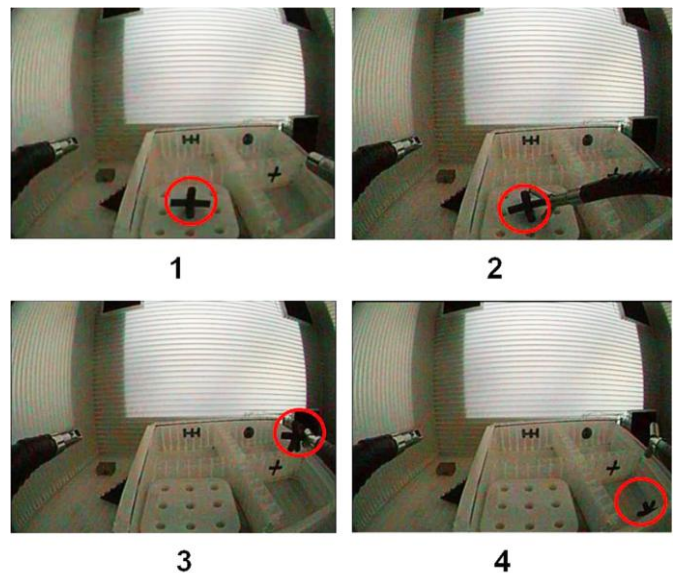

**Figure 5.** Image illustrations of each of the four steps, described verbally in Table 1, of the *pick-and-drop task* when performed with the right hand. At the beginning of each session, the object was replaced in the departure box at the same position and with the same orientation (see image 1, top left).

Once the distal tool correctly positioned, the object has to be grasped with the tool tips (*step 2*). Once firmly held by the gripper, the object has to be moved laterally to a position on top of the target box (*step 3*) with the distal extremity of the tool in the correct position for dropping the object into the target box without missing (*step 4*). A user starts and ends a given task session by pushing a button that is wirelessly connected to the computer receiving the data. The robot-assisted precision grip task involves positioning the instrument tip through movements from left to right, from up to down, and from forward to backwards. Manipulation of one instrument is required to perform this task here. Given the limited degrees of freedom for hand and finger movements, the user is bound to perform the task with each hand in rather the same way. The locally deployed grip forces will vary depending on how skilled a user has become in performing the task. Here, grip force data collected from two users with distinct levels of task skill (expert *vs* novice) were analyzed. The hand preference profiles of these two users were different. The expert user has been practicing on the system since its manufacture and become a highly proficient user, with years of user experience and more than 100 hours of training in this specific task. This expert is proficient in using the system with his dominant (left) hand, and has only a moderate preference for using his left hand in every day manual tasks. The other user was a complete novice who had never used the system before, nor had he any prior experience with any similar system, with a strong preference for right hand use in every day manual tasks. He was given one hour to familiarize himself with the system with both hands prior to the experiments. Their hand sizes were about the same, and the sensor gloves were developed specifically to fit the hands of average-proportioned male individuals. Expertise in this type of task is consistently reflected by specific performance parameters such as the average task time taken in a session, and the number of incidents, accounting for the number of times an object is dropped, and the number of unsuccessful grasp attempts or tool-trajectory readjustments. Each user performed the task with the dominant and the non-dominant hand in ten successive sessions for each hand without major breaks between sessions.

*2.5 Statistical analyses*

The variability in the novice and the expert data for each hand across the ten sessions was statistically assessed for all sensors that produced meaningful output. The goal of this exercise is to bring to the fore skill-specific differences in grip-force variability across sensors and sessions. This was achieved by computing and plotting the variance in the data in terms of standard deviations from the mean, which express the squared differences between the observations and the group mean divided by the number of data points for each group of observations. The statistical analyses (2-Way ANOVA) between the raw grip force data from a functionally relevant subset (Table 2) of three of the twelve sensors (S5, S6, S7) for the expert and the novice from their first and last task sessions with the dominant hand was performed using the MATLAB toolbox. These statistical analyses were designed to test for functionally specific effects, and their probability limits, of the 'Expertise' factor (expert *vs* novice) and the 'Session' factor (first session *vs* last session).

**Table 2.** Functionally relevant sensors (S5, S6, S7) for the controlled manipulation of the cylindrical robotic handles. Output measures for these sensors from the expert's and the novice's first and last task sessions with the dominant hand were compared to bring to the fore proficiency-specific differences.

| Sensor | Finger | Grip Force Control |
|:---:|:---:|:---:|
| **S5** | middle | gross grip force deployment |
| **S6** | ring | non-specific grip force support |
| **S7** | pinky | precision grip control |

These same selected data were then submitted to spatiotemporal analysis in terms of Average peak amplitudes in milliVolt (*AmV*) for successive temporal windows of a fixed size of 2000 msec each in the given individual sessions. With one signal recorded every 20 msec and 100 signals per time window of 2000 msec, we have *AmV=mVtotal/100*, which is the total sum of *mV* recorded in a time window given divided by the total number of signals in that time window. Since expert and novice had different hand use preferences in everyday life, as explained here above in *2.4*, the differences shown for the dominant hand are likely to best reflect specific characteristics relating to the skill level in this task here.

*2. 6 Neural network model*

A neural network architecture described in detail in previous work [56-59] referring to functional properties of the Quantization Error (QE) in the output of a Self-Organizing Map (SOM) was exploited for modeling the variability in the whole data set of grip forces recorded from ten sensors and ten successive task sessions with the dominant and non-dominant hands of the expert and the novice. The SOM is described formally as a nonlinear, ordered, smooth mapping of high-dimensional input data onto the elements of a regular, low-dimensional array. The input variables are defined here as a real vector $\boldsymbol{x}$ of n dimensions (the input representation), and each element therein is associated with a parametric real vector $\boldsymbol{m_i}$ of n dimensions (the model representation). Assuming a general distance measure between $x$ and $m_i$, denoted by d($x$, $m_i$), the map of an input vector $x$ on the SOM array is defined as the array element $m_c$ that best matches $x$ (smallest d($x$, $m_i$)). During the learning process, models topographically close in the map up to a certain geometric distance, denoted by $h_{ci}$, will activate each other to learn something from their shared input $x$. This then results in a local relaxation, or smoothing effect, on the models in this neighborhood, which in continued learning leads to global ordering. SOM learning is represented by the equation

$$m(t+1) = m_i(t) + \alpha(t)\, h_{ci}(t)\, [x(t) - m_i(t)] \qquad (2)$$

where t =1,2,3...is an integer, the discrete-time coordinate, $h_{ci}(t)$ the neighborhood function (a smoothing kernel defined over the map points which converges towards zero with time) and $\alpha(t)$ is the learning rate. At the end of the *winner-take-all* learning process, each input vector $x$ becomes associated with, or mapped to, its best matching model. The difference between $x$ and $m_c$, $\|x - m_c\|$, is reflected by the quantization error *QE*. The *QE* of $x$ is given by

$$QE = 1/N \sum_{i=1}^{N} \|x_i - m_{c_i}\| \qquad (3)$$

where *N* is the number of input vectors $x$.

*2. 7 Rationale for the neural network architecture*

The SOM implemented here was designed to map a brain inspired mechanoreceptor-to-brain model network in terms of a 7 by 7 map generating a fully connected network of 49 neurons where each of the sensors for which data were exploited contributes to the final synaptic weight of each neuron. The *QE* in the SOM output, the *SOM-QE* [56,57], expresses a difference between an input representation and its model representation at a given moment in time *t*, and captures variations of this difference with time [59,60]. The *QE* in the output of the brain inspired neural network map reflects the amount of variability in the grip forces of the expert and the novice at any given moment in time, and the evolution of this variability with time, i.e. across the ten successive task sessions. Intra- and inter-individual grip force variability directly translate skill levels [39], and tend to decrease with practice, converging towards an optimum characteristic of expert performance.

**3. Results**

The results from the sensor calibration procedure, the neural network modeling of skill specific grip force variability exploiting all the sensor data that produced significant output, and the individual spatiotemporal grip force profiles for expert and novice dominant hand data for selected sensors with a particular functional significance (Table 2) in the precision grip task are presented here below.

*3.1. Sensor calibration*

The relationship between force (g) and tension (mV) from the sensor calibrations, explained under *2.2.* of the Materials and Methods section, was analyzed. The raw sensor output data in the experiments here vary essentially within the linear range of this relationship (Fig.6). All the following analyses and comparisons are therefore computed directly on the voltage levels at the millivolt (mV) scale.

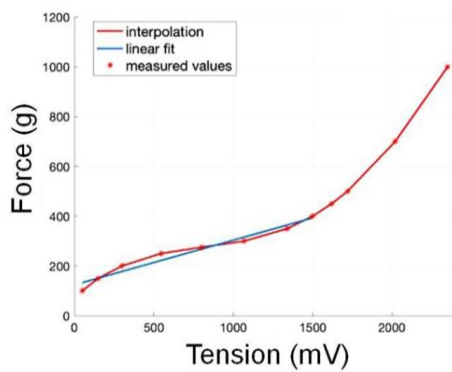

**Figure 6.** Relationship between force (in gram) and tension (in milliVolt) of the sensor output. The sensor output measured in the experiments varies essentially within the linear range (blue line).

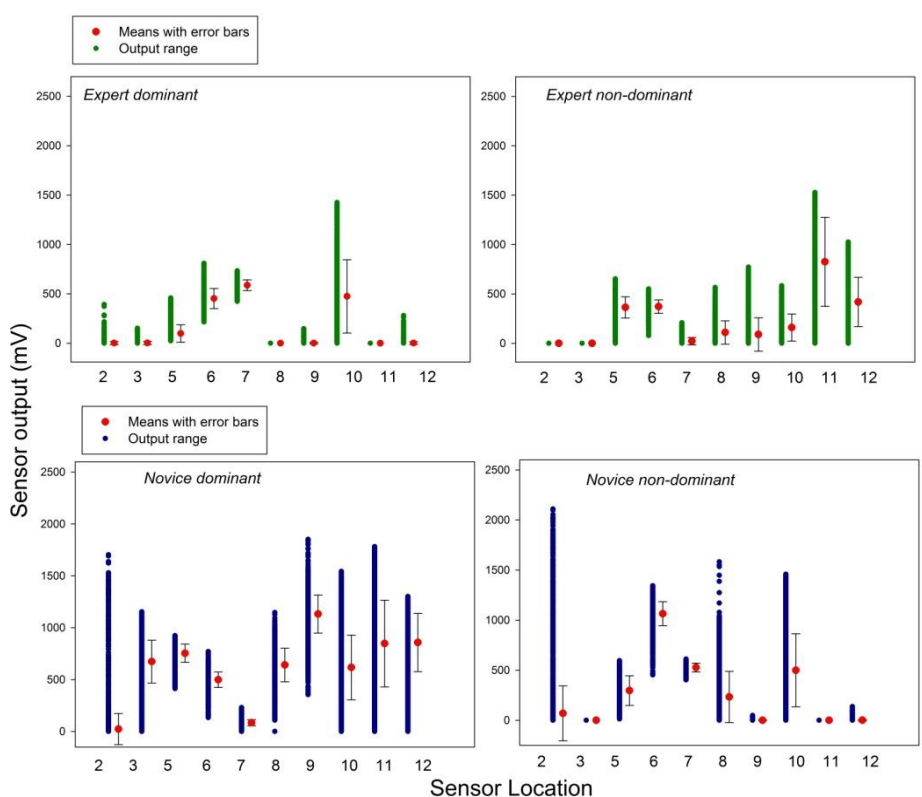

**Figure 7.** Sensor output range, with means across ten sessions and their standard errors for the expert and the novice performing with the dominant and the non-dominant hand.

*3.2. Skill-specific grip force variability*

The numerical range of observations (in mV) was determined for each sensor across sessions (ten per hand), for expert and novice performing with the dominant and non-dominant hand. The results of this descriptive analysis are shown here above (Fig. 7). The variability in grip forces deployed by the novice and the expert by the dominant and the non-dominant hands was then analyzed on the basis of the raw data from each individual session. Temporal variability in biosensor data expresses the evolution of the amount of functional "noise" in the living system under study, and was determined here in terms of standard deviations (STD) from the mean (not standard errors of the mean as shown in Fig.7). STD express the squared differences between observations (measurements) and the group mean (for a given individual session here), divided by the number of data points in the group. These were plotted as a function of the session order for all sensors that produced meaningful output (Fig 8.).

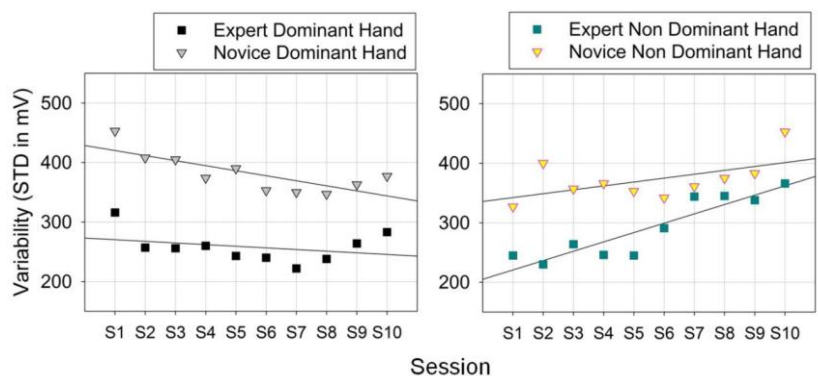

**Figure 8:** Variability (STandard Deviations of the means) in grip forces deployed by the novice and the expert by the dominant and the non-dominant hands in ten successive task sessions.

An increase in variability of grip forces (STD) with time (sessions) during task performance with the non-dominant hand is observed in both users (Fig. 8, right). It is therefore not skill-related, but may indicate a tendency to fatigue in the non-preferred hand when the number of repeated task sessions increases. This interpretation holds ground given the difference in forearm muscle size between dominant and non-dominant extremities [60].

*3.3. Neural network model*

The neural network model (SOM) described under *2.6* was then run on the same variability data for each user type and session. The QE in the output of these analyses is plotted here below as a function of the session order (Fig. 9). Further statistical analyses of the grip force and the SOM model data (t-test paired comparisons) yield significant effects of task expertise on the STD ($t(1,18)=22.34$; $p<.001$ for the dominant hand; $t(1,18)=7.43$; $p<001$ for the non-dominant hand). These are mirrored by significant effects on the SOM-QE ($t(1,18)=9.27$; $p<.001$ for the dominant hand; $t(1,18)=4.09$; $p<.001$ for the non-dominant hand). These result show that task skill evolution reflected by grip force variability as a function of time, in an expert by comparison with a novice, can be reliably predicted by unsupervised learning in an artificial neural network map mimicking functional properties of a biological receptor network in the somatosensory brain. These networks develop task specific functional synergies aimed at reducing motor redundancy [43]. In our previous work [40], correlation analyses revealed skill-specific differences in co-variation patterns in individual grip force profiles reflected by an optimum of significant and functionally specific co-variation of a few sensors in the expert dominant hand, and non-specific co-variation of a large amount of sensors in the dominant hand of novices.

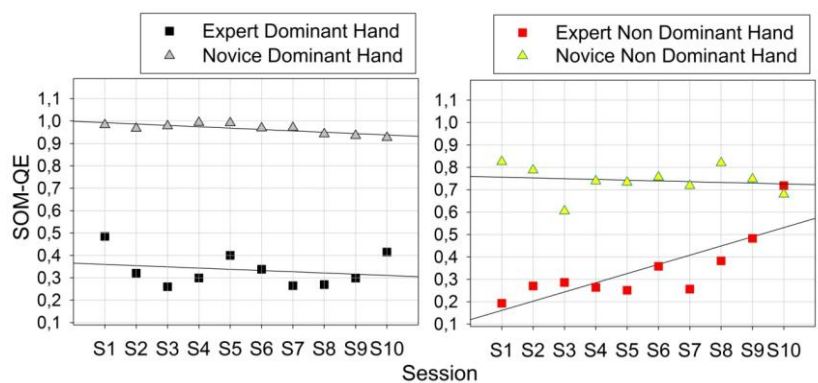

**Figure 9:** *QE* in the output of the brain inspired neural network model (SOM) run on the grip force data from the different sessions; the difference in grip force variability between the expert and the novice is reliably predicted by the neural network map output metric.

*3.3. Functionally motivated spatiotemporal profiling*

To highlight functionally specific task-relevant grip force differences in the sensor profiles of expert and novice, the following spatio-temporal analysis zooms in on a selected number of functionally relevant sensors in the expert's and the novices dominant hands. In [40] we had shown that robotic task skill is reflected by an optimum of significant positive correlations between output of dominant-hand sensors with similar function (S5 and S6) and significant negative correlations between dominant-hand sensors with different function (S6 and S7, for example). Conversely, a novice profile exhibits non-specific co-variation of a large amount of sensors in both hands, which translates functional redundancy characteristic of unskilled manipulation [43,45]. Individual temporal grip force profiles from sensors S5, S6, and S7 were submitted to spatio-temporal analysis. This was achieved by computing the Average peak amplitudes in milliVolt (*AmV*) for fixed successive temporal windows of 2000 millliseconds (msec) from a given individual session, as explained in *2.5* here above. These profiles, comparing between the first and the last individual sessions of the expert and the novice, are shown here below (Fig. 10). Statistical comparison (2-Way ANOVA) between the original raw data of expert and novice from their first and last task sessions reveal significant interactions between 'expertise' (2 factor levels) and 'session' (2 factor levels) for all three sensors considered here (S5, S6, S7). For sensor S5 on the middle finger (gross grip force deployment), the mean (*m*) grip forces and their standard errors (*sem*) from the first session yield *m=241mV /sem=4.3* for the expert and *m=790mV/sem=2.7* for the novice, showing that the latter deploys about three times as much unnecessary gross grip force by comparison with the expert. This expertise-specific difference in proportional gross grip force deployed by the middle finger is even larger in the last session, with *m=78mV/sem=4.9* for the expert, and *m=640mV/sem=3.6* for the novice. The interaction between the 'expertise' and 'session' factors for sensor S5 is significant with $F(1,2880)=28.65$; $p<.001$. For sensor S6 on the ring finger, which has no particular role in grip force control, the differences between the grip force profiles of novice and expert are minimal, as would be expected, in the first session with *m=576mV /sem=3.8* for the expert and *m=504mV/sem=2.4* for the novice, and in the last session with *m=474mV /sem=4.5* for the expert and *m=445mV/sem=3.3* for the novice. The interaction between the 'expertise' and 'session' factors for sensor S6 is, however, significant with $F(1,2880)=35.86$; $p<.001$, which is explained by the fact that grip forces, i.e. amplitudes in *mV*, diminish in both users from the first to the last session, but not by the same amounts. For sensor S7 on the small finger (critically important for fine grip force control), the expertise-specific difference between the two user profiles is characterized by the novice deploying largely insufficient grip forces, from the first session with with *m=98mV /sem=1.2* to the last with *m=78mV/sem=1.6,* while the expert produces sufficient grip force for fine movement control from the first session with *m=594mV /sem=1.8* to the last with *m=609mV/sem=2.2.*

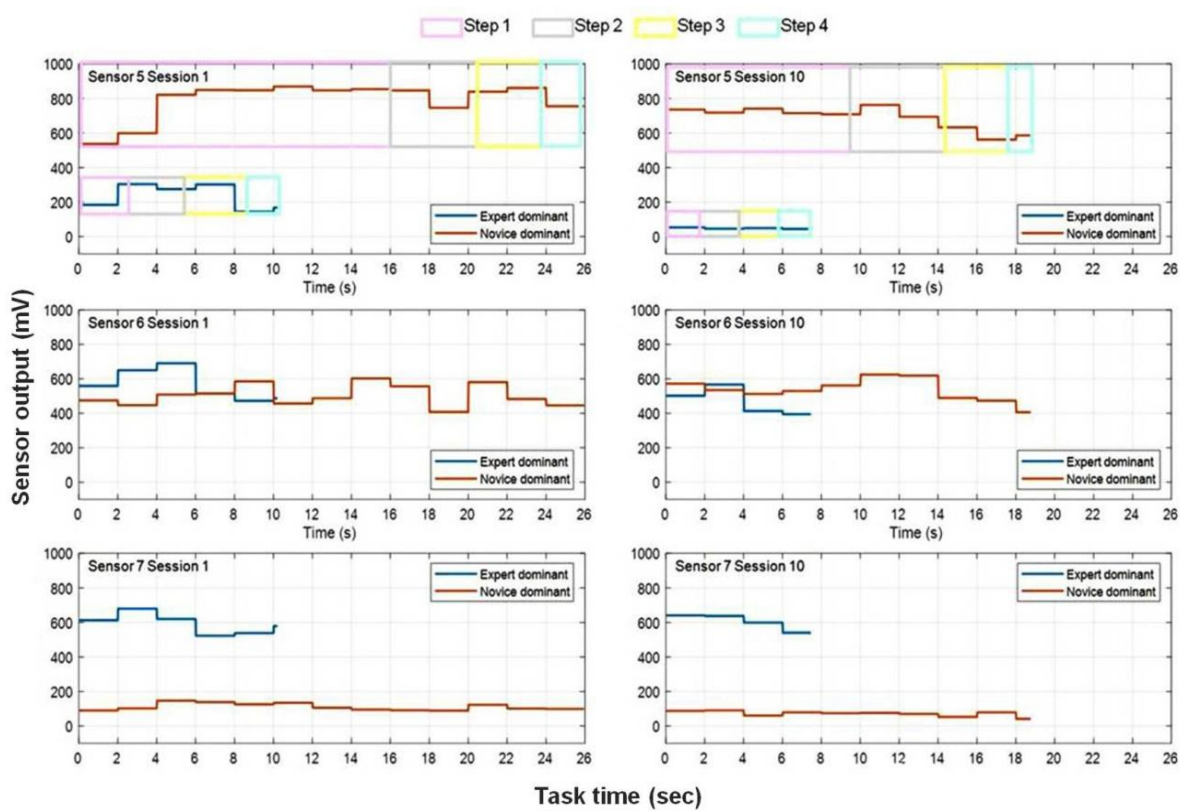

**Figure 10.** Individual grip force profiles showing average peak amplitudes (mV) from sensors 5, 6, and 7 for fixed successive temporal windows of 2000 milliseconds in a given session, for the first and last of ten sessions of the expert and the novice. Relative durations of each of the four critical task steps within a given session are highlighted by the colored boxes corresponding to those shown in Table 2 on the example of sensor 5 (top left and top right).

The interaction between the 'expertise' and 'session' factors for sensor S7 is highly significant with $F(1,2880)=188.53$; $p<.001$. Average task times across sessions and hands are considerably shorter for the expert (Table 3), with 10.2 sec across hands in the first session and 7.5 sec across hands in the last indicating a minor practice effect. The novice takes more than twice as long (~25 sec) in the first session compared with the expert, with a 30% time gain in the last session (18.8 sec), indicating a temporal training effect. Regarding incidents (trajectory adjustments, grip failures, drop misses) across all sessions, the task videos reveal a total number of 20 in the novice across sessions and hands, and only three small trajectory adjustments in the expert's last three sessions with the non-dominant hand (Table 3). Incidents during task completion directly impact on the completion times. Their effect on grip force depends on the type of incident. While the minor tool trajectory corrections (cf. expert data) may be deemed to have little effect on magnitude and variability of grip force, the major incidents observed in all sessions of the novice most likely affect both [8].

**Table 3:** Average times taken to complete the task with the dominant and non-dominant hand in the first and last of ten sessions per hand. The total number of incidents across all sessions are shown in the last column.

| Skill Level | 1st session duration | Last Session | Incidents |
|---|---|---|---|
| Expert | 10.20 | 7.48 | 3 |
| Novice | 24.56 | 18.78 | 20 |

## 4. Discussion

The analyses reveal expertise-specific differences in the spatiotemporal grip force profiles of an expert and a novice repeatedly performing a 2D image-guided robot-assisted precision task. These differences are evaluated here with regard to their functional implications, under the light of previous work on the role of finger grip forces and prehensile synergies, which are centrally controlled in the human brain for human motor performance and control. Skill-related differences in grip force deployment here are reflected by a larger general grip force variability and higher grip force magnitude across hands and sessions in the novice. One of the more particular aspects of task proficiency in this study context here concerns the proportional gross grip force deployed by the middle finger of the dominant hand. While the novice deploys too much of it, the expert has learnt to optimize and deploy it parsimoniously, as shown by the detailed analyses here. Excessive grip force deployment appears common in novice surgeons in robotic surgery when there is absence of haptic feed-back [50]. This may be corrected at the earliest stages of training through verbal feed-back. Since grip force modulation is under higher-level cortical control [9,33,43,69] involving the frontal lobe, raising awareness in the novice verbally could promote faster adaptive learning. Another functionally important aspect concerns the precision grip force control of the cylindrical handles by the small finger, critically important in surgical and other precision tasks. The difference between the two users here is characterized by the novice deploying insufficient small finger grip forces with the preferred (dominant) hand, with no major evolution between the first and the last task sessions. When looking at the grip forces deployed by the ring finger, which plays no meaningful role in grip force control when manipulating cylindrical objects, differences between the profiles of novice and expert are minimal, as would be expected, and do not evolve much across sessions. Sensors were positioned in the glove to optimally fit the cylindrical handles of the robotic control device. The functionally relevant subset in the spatiotemporal analyses were selected in this specific context on the basis of findings from previous studies on finger-specific functional implications for manipulation of objects with variable shape properties [10,11,12,16]. Previous work [16] has shown that force distributions for cylindrical grips differ between dominant and non-dominant hand in healthy subjects, as shown in this study here. For control handles with other shapes (elliptic, circular etc.), finger and hand specific functional synergies may be different. These can then be benchmarked using the grip force profiles of highly proficient experts. The synergy of finger grip function is self-organizing and highly plastic. The complex anatomy of the human hand allows for a large number of postures and finger combinations to attain optimal grip force synergies in tasks with different constraints and the manipulation of objects with different shapes [42,43,44]. The total grip force magnitude, or mV amplitude, tends to diminish across sessions with practice in both users here. Regarding task times, the novice takes more than twice as long performing the precision task by comparison with the expert, but at the end scores a 30% time gain indicating a considerable temporal training effect, especially in the first critical task step, which is the most difficult to perform under 2D image guidance given the absence of veridical depth cues in the camera image. Moving the tool along a virtual trajectory towards the object location requires movement away from the body in the surgeon's peripersonal space and is difficult to scale under conditions of 2D image input. It implies compensating for physically missing depth information by slowing down, which not only results in longer task times but also in less precise tool-movements [51-55]. This explains why the largest training gain in total task time of the novice is observed for task step 1. The 2D camera of the robotic system here represents a limitation. Using 3D camera systems, as recently demonstrated in laparoscopic approaches [61], may help overcome this problem. The grip force analyses shown here in this work can be performed in real-time to monitor manual/bimanual precision tasks, control performance quality, or to prevent risks in surgery systems where excessive grip forces can directly cause tissue damage [51]. Task skill-related variations in grip forces are reliably predicted, as illustrated here, by the output metric of the brain-inspired neural network architecture simulating functional properties of somatosensory feed-back circuitry in the human brain. Combining grip force sensor technology with predictive modeling by computationally parsimonious Artificial Intelligence promises functionally meaningful and economic automated analysis of surgical task skill evolution. Current state of the art in robotic assistance [62,63] for surgical procedures with unified master-slave control systems [64,65] has a considerable potential for augmenting the precision and technical capability of physicians, but some challenges still need to be met in terms of optimized system architecture, software, mechanical design, imaging systems, and user interface design and management for maximum safety. To avoid single observer bias [66], objective quantitative performance criteria need to be worked out for defining gold standards of true expert performance in this emerging realm of assistive technology, pushing optimal training programs for novices. Cogently designed and parsimoniously deployed Artificial Intelligence [20,67] can help move things forward in this direction. Finally, the control of the human hand by the brain has evolved as a function of environmental constraints in interaction with the other sensory systems, and grip force profiles are a direct reflection of the complex cognitive and behavioral synergies these interactions have produced. Sensory cues provided by somatosensation, vision, hearing, and smell play an important role in grip force scaling [1]. When interacting with objects of uncertain properties providing insufficiently reliable somatosensory feed-back, individuals use somatosensory memory representations from previous trials to plan grip forces [68], and patients with massive somatosensory loss can still scale and time grip forces and adjust them across different object handling tasks on the basis of memory-based, anticipatory and online control processes to compensate for the loss of somatosensory feedback [69]. The range of possibilities offered by wearable wireless sensor technology in the study of human cognitive processing extends well beyond the field of human-robot interaction.

## 5. Conclusions

Wearable wireless sensor technology has permitted exploring grip forces deployed for grasping, lifting and manipulating objects under conditions of variable external constraints and sensory input in novel tasks that require the rescaling of perceptual responses to obtain behavioral success. Profiling the grip forces of individuals with variable skill levels in image-guided tasks that require to interact with a robotic device reveals some of the dynamic functional changes that take place in the brain during practice and learning. Human-robot interaction represents unprecedented challenges for perceptual and motor adaptation in environmental contexts of high sensory uncertainty. The insights from this study here may contribute towards improving the outcome of new types of surgery, in particular the single-port approaches such as NOTES (Natural Orifice Transluminal Endoscopic Surgery) and SILS (Single Incision Laparoscopic Surgery). Beyond this particular context, grip force analysis offers various perspectives for studying cognitive processes in larger realm, relating to multisensory interactions with hearing [29] or vision [31]. Successful grip force deployment involves central processes of neural control [32,33], and grip force is currently explored as a marker of brain health [70] in clinical studies of cognitive disorders such as major chronic depression [71], Parkinson's disease [72], or the non-pathological cognitive decline in ageing [73-76]. As a directly measurable behavioral correlate of self-organizing control mechanisms in brain learning [77], grip force patterns and their evolution are suited for feeding theoretical approaches and hypotheses that exploit neural network architectures driven by unsupervised biological learning [20].

**Author Contributions:** Conceptualization, B.D.L., R.L., J. W., M. de M.; methodology, B.D.L., F.N. M. de M., J. W.; software, P.Z.; validation, R. L., J.W., F.N., P.Z. and B.D.L.; formal analysis, B.D.L., R.L., J.W.; investigation, B.D.L., R. L., J. W.; resources, F. N., P.Z.; data curation, F.N, B.D.L., R. L., J. W.; writing—original draft preparation, B.D.L.; writing—review and editing, B.D.L., R.L., J. W., F. N., M. de M.

**Funding:** This research received funding from the Initiative D'EXcellence (IDEX) de l'Université de Strasbourg.

**Acknowledgments:** Material support from CNRS is gratefully acknowledged.

**Conflicts of Interest:** The authors declare no conflict of interest.